\documentclass{article}

\usepackage{aaai16}

\usepackage[T1]{fontenc} 
\usepackage[scaled=0.2]{beramono}
\usepackage{graphicx}
\usepackage{times}
\begin{document}

\title{Reuse of Neural Modules for General Video Game Playing}

\author{
Alexander Braylan \hspace{15pt}
Mark Hollenbeck \hspace{15pt}
Elliot Meyerson \hspace{15pt}
Risto Miikkulainen\\
Department of Computer Science,
The University of Texas at Austin\\
\texttt{\{braylan,mhollen,ekm,risto\}@cs.utexas.edu}}

\maketitle

\begin{abstract}
A general approach to knowledge transfer is introduced in which an agent controlled by a neural network adapts how it reuses existing networks as it learns in a new domain. Networks trained for a new domain can improve their performance by routing activation selectively through previously learned neural structure, regardless of how or for what it was learned. A neuroevolution implementation of this approach is presented with application to high-dimensional sequential decision-making domains. This approach is more general than previous approaches to neural transfer for reinforcement learning. It is domain-agnostic and requires no prior assumptions about the nature of task relatedness or mappings. The method is analyzed in a stochastic version of the Arcade Learning Environment, demonstrating that it improves performance in some of the more complex Atari 2600 games, and that the success of transfer can be predicted based on a high-level characterization of game dynamics. 
\end{abstract}

\section{Introduction}

The ability to apply available previously learned knowledge to new tasks is a hallmark of general intelligence. \emph{Transfer learning} is the process of reusing knowledge from previously learned \emph{source} tasks to bootstrap learning of \emph{target} tasks. In long-range sequential control domains, such as robotics and video game-playing, transfer is particularly important, because previous experience can help agents explore new environments efficiently \cite{Taylor09,Konidaris12}. Knowledge acquired during previous tasks also contains information about an agent's domain-independent decision making and learning dynamics, and thus can be useful even if the domains seem unrelated. 

Existing approaches to transfer learning in such domains have demonstrated successful transfer of varying kinds of knowledge, but they make two fundamental assumptions that restrict their generality: (1) some sort of a priori human-defined understanding of how tasks are related, and (2) separability of knowledge extraction and target learning. The first assumption limits how well the approach can be applied by restricting its use only to cases where the agent has been provided with this additional relational knowledge, or, if it can be learned \cite{Talvitie07,Taylor08,Ammar15a}, cases where task mappings are useful. The second assumption implies that it is known what knowledge will be useful and how it should be incorporated \emph{before} learning on the target task begins, preventing the agent from adapting the way it uses source knowledge as it gains information about the target domain. 

General ReUse of Static Modules (GRUSM) is proposed in this paper as a general neural network approach to transfer learning that avoids both of these assumptions. GRUSM augments the learning process to allow learning networks to route through existing neural modules (source networks) selectively as they simultaneously develop new structure for the target task. Unlike previous work, which has dealt with mapping task variables between source and target, GRUSM is domain-independent, in that no knowledge about the structure of the source domain or even knowledge about where the network came from is required for it to be reused. Instead of using mappings between task-spaces to facilitate transfer, it searches directly for mappings in the solution space, that is, connections between existing source networks and the target network. This approach is motivated by studies that have shown in both naturally occurring complex networks \cite{Milo02} and in artificial neural networks \cite{Swarup06} that certain network structures repeat and can be useful across domains, without any context for how exactly this structure should be used. This work is further motivated by the idea that neural resources in the human brain are reused for countless purposes in varying complex ways \cite{Anderson10}. 

In this paper, an implementation of GRUSM based on the Enforced Subpopulations (ESP) neuroevolution framework \cite{Gomez97,Gomez99} is presented. The approach is validated on the stochastic Atari 2600 general game playing platform, finding that GRUSM-ESP improves learning for more complex target games, and that these improvements may be predicted based on domain complexity features. This result demonstrates that even without traditional transfer learning assumptions, successful knowledge transfer via general reuse of existing neural modules is possible and useful for long-range sequential control tasks. In principle, this approach scales naturally to transfer from an arbitrary number of source tasks, which suggests that in the future it may be possible to build GRUSM agents that accumulate and reuse knowledge throughout their lifetimes across a variety of diverse domains. 


\section{Background}
\label{SectionBackground}

Transfer learning encompasses machine learning techniques that involve reusing existing \emph{source} knowledge in a different \emph{target} task or domain. A \emph{domain} is an environment in which learning takes place, characterized by the input and output space; a \emph{task} is a particular function from input to output to be learned \cite{Pan10}. In sequential-decision domains, a task is characterized by the values of sensory-action sequences corresponding to the pursuit of a given goal. A taxonomy of types of knowledge that may be transferred was also enumerated by Pan and Yang. Because the GRUSM approach reuses the structure of existing neural networks, it falls under \emph{feature representation transfer}.

\subsection{Transfer Learning for RL}

Transfer learning for sequential decision-making domains has been studied extensively within the \emph{reinforcement learning} (RL) paradigm \cite{Taylor09}. Reinforcement learning domains are often formulated as Markov decision processes (MDPs) in which the state space comprises all possible observations, and the probability of an observation depends only on the previous observation and action taken by a learning agent. However, many real world RL domains are non-Markovian, including many Atari 2600 games, for example, the velocity of a moving object cannot be determined by looking at a single frame. 

The Atari 2600 platform also supports a wide variety of games. Existing RL approaches to transfer differ on the types of differences allowed between source and target task. Some approaches that are general with respect to the kind of knowledge that can be transferred are restricted in that they require a consistent \emph{agent-space} \cite{Konidaris12}, or an a priori specification of inter-task \emph{mappings} defining relationships between source and target state and action variables \cite{Brys15}. Existing approaches to transfer learning that encode policies as neural networks require such a specification \cite{Taylor07a,Verbancsics10}. On the other hand, existing modular neuroevolution approaches that are more general with respect to connectivity \cite{Reisinger04,Khare05} have not been applied to cross-domain transfer. 

Some of the most general existing approaches to transfer for RL automatically learn task mappings, so they need not be provided beforehand. These approaches are general enough to apply to any reinforcement learning domains, but initial approaches \cite{Taylor08,Talvitie07} were intractable for high dimensional state and action spaces due to combinatorial blowup in the number of possible mappings. However, recent approaches in policy gradient RL \cite{Ammar15a,Ammar15b} can both tractably learn mappings and be applied across diverse domains. These approaches have been successful in continuous control domains, but it is unclear how they would scale to domains with many discretely-valued features such as Atari. Also, the above approaches assume MDP environments, whereas GRUSM can use recurrent neural networks to extend to POMDPs.

\subsection{General Neural Structure Transfer}
There are existing algorithms similar to GRUSM in that they make it possible to reuse existing neural structure. They can apply to a wide range of domains and tasks in that they automatically select source knowledge and avoid inter-task mappings. For example, \citeauthor{Shultz01} (\citeyear{Shultz01}) developed a technique to build increasingly complex networks by inserting source networks chosen by how much they reduce error. This technique is only applicable to supervised learning, because the source selection depends heavily on an immediate error calculation. Also, connectivity between source and target networks is limited to the input and output layer of the source. As another example, \citeauthor{Swarup06} (\citeyear{Swarup06}) introduced an approach that creates sparse networks out of primitives, or commonly used sub-networks, mined from a library of source networks. This subgraph mining approach depends on a computationally expensive graph mining algorithm, and tends to favor exploitation over innovation and small primitives rather than larger networks as sources. 

The GRUSM approach is more general in that it can be applied to unsupervised and reinforcement learning tasks, makes few a priori assumptions about what kind of sources and mappings should work best, and is able to develop memory via recurrent connections. Although an evolutionary approach is developed in this paper, GRUSM should be extensible to any neural network-based learning algorithm. 

\section{Approach}
\label{SectionApproach}

This section introduces the general idea behind GRUSM, provides an overview of the ESP neuroevolution framework, and describe the particular implementation: GRUSM-ESP.

\subsection{General ReUse of Static Modules (GRUSM)}

The underlying idea is that an agent learning a neural network for a target task can reuse knowledge selectively from existing neural modules (source networks) while simultaneously developing new structure unique to a target task. This approach attempts to balance reuse and innovation in an integrated architecture. Both source networks and new hidden nodes are termed \emph{recruits}. Recruits are added to the target network during the learning process. Recruits are incorporated adaptively into the target network as it learns connection parameters from the target to the recruit and from the recruit to the target. All internal structure of source networks is \emph{frozen} to allow learning of connection parameters to remain consistent across recruits. This mechanism forces the target network to transfer learned knowledge, rather than simply overwrite it. Connections to and from source networks can, in the most general case, connect to any nodes in the source and target, minimizing assumptions about what knowledge will be useful.

A GRUSM network is a 3-tuple $\mathcal{G} = (M,S,T)$ where $M$ is a traditional neural network (feedforward or recurrent) containing the new nodes and connections unique to the target task, with input and output nodes corresponding to inputs and outputs defined by the target domain; $S$ is a (possibly empty) set of pointers to recruited source networks $\mathcal{S}_1,...,\mathcal{S}_k$; and $T$ is a set of weighted \emph{transfer connections} between nodes in $M$ and nodes in source networks, that is, for any connection $((u,v),w) \in T$, $(u \in M \wedge v \in \mathcal{S}_i) \vee (u \in \mathcal{S}_i \wedge v \in M)$ for some $0 \leq i \leq k$. This construction strictly extends traditional neural networks so that each $\mathcal{S}_i$ can be a traditional neural network or a GRUSM network of its own. When $\mathcal{G}$ is evaluated, only the network induced by directed paths from inputs of $M$ to outputs of $M$, including those which pass through some $\mathcal{S}_i$ via connections in $T$ is evaluated. During each evaluation of $\mathcal{G}$, all recruited source network inputs are fixed at 0, since the agent is concerned only with performing the current target task. The parameters to be learned are the usual parameters of $M$, along with the contents of $S$ and $T$. The internal parameters of each  $\mathcal{S}_i$ are \emph{frozen} in that they cannot be rewritten through $\mathcal{G}$. 

The motivation for this architecture is that if the solution to a source task contains \emph{any} information relevant to solving a target task, then the neural network constructed for the source task will contain \emph{some} structure (subnetwork or module) that will be useful for a target network. This has been previously observed in naturally occurring complex networks \cite{Milo02}, as well as cross-domain artificial neural networks \cite{Swarup06}. Unlike the subgraph-mining approach to neural structure transfer \cite{Swarup06}, this general formalism makes no assumptions as to what subnetworks actually will be useful.  One interpretation is that a lifelong learning agent maintains a system of interconnected neural modules that it can potentially reuse at any time for a new task. Even if existing modules are unlabeled, they may still be useful, due to the fact that they contain knowledge of how the agent can successfully learn. Furthermore, advances in reservoir computing \cite{Luko09} have demonstrated the power of using large amounts of frozen neural structure to facilitate learning of complex and chaotic tasks.

The above formalism is general enough to allow for an arbitrary number of source networks and arbitrary connectivity between source and target. In this paper, to validate the approach and simplify analysis, at most one source network is used at a time and only connections from target input to source hidden layer and source output layer to target output are permitted. By not allowing target input to connect to source input, this restriction avoids high-dimensional transformations between domain-specific sensor substrates, and more intuitively captures the domain-agnostic goals of the approach, differentiating the approach from previous methods that have used direct mappings between sensor spaces. This restriction is sufficient to show that the implementation can reuse hidden source features successfully, and it is possible to analyze the cases in which transfer is most useful. Future refinements are discussed in the Discussion and Future Work section. The current implementation, described below, is a neuroevolution approach based on ESP.

\subsection{Enforced Subpopulations (ESP)}
Enforced Sub-Populations (ESP; \citeauthor{Gomez97} \citeyear{Gomez97}; \citeyear{Gomez99}) is a neuroevolution technique in which different components of a neural network are evolved in separate \textit{subpopulations} rather than evolving the whole network in a single population. ESP has been shown to perform well in a variety of reinforcement learning domains, and has shown promise in extending to POMDP environments, in which use of recurrent connections for memory is critical \cite{Gomez99,Gomez05,Schmidhuber07}. In traditional ESP, there is a single hidden layer, each neuron of which is evolved in its own subpopulation. Recombination occurs only between members of the same subpopulation, and mutants in a subpopulation derive only from members of that subpopulation. The genome of each individual in a subpopulation is a vector of weights corresponding to the weights of connections to and from that neuron, including node bias. In each generation, networks to be evaluated are randomly constructed by inserting one neuron from each subpopulation. Each individual that participated in the network receives the fitness achieved by that network. 

When fitness converges, i.e., does not improve over several consecutive generations, ESP  makes use of \emph{burst phases}. In initial burst phases each subpopulation is repopulated by mutations of the single best neuron ever occuring in that subpopulation, so that it reverts to searching a $\delta$-neighborhood around the best solution found so far. If a second consecutive burst phase is reached, i.e., no improvements were made since the previous burst phase, a new neuron with a new subpopulation may be added \cite{GomezThesis}.

\subsection{GRUSM-ESP}

\begin{figure}[t]
\centering
\includegraphics[width=0.465\textwidth]{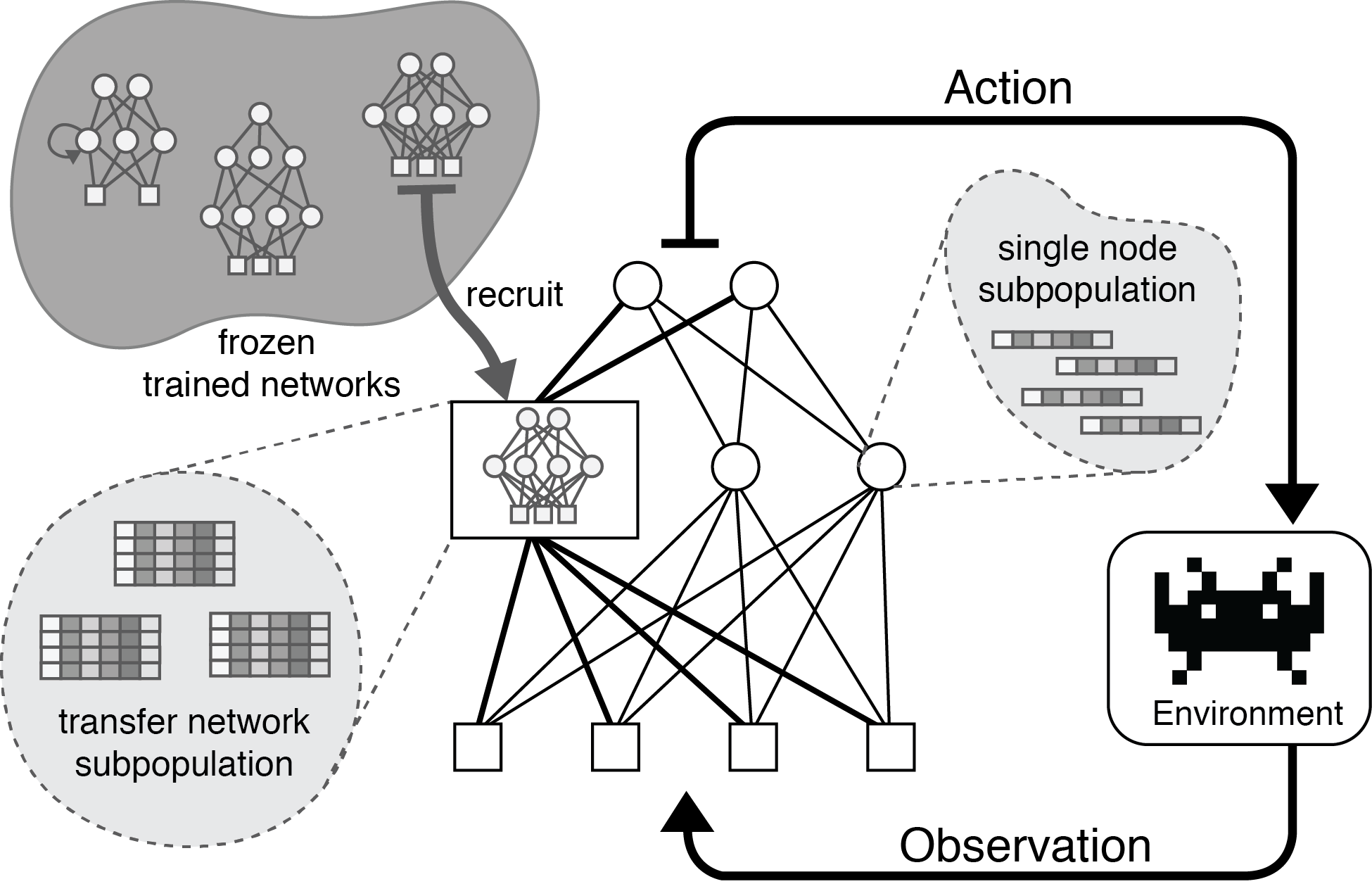}
\caption{\label{FigureGESP} The GRUSM-ESP architecture, showing the balance between reused and new structure. In this example, the target network has three recruits: one source network, and two single nodes. Each bold edge between target network nodes and source network recruit indicate connections to multiple source nodes. The genome in each subpopulation encodes weight information for the connections from and to the corresponding recruit.}
\end{figure}

The idea of enforced sub-populations is extended to transfer learning via GRUSM networks. For each reused source network $\mathcal{S}_i$, the transfer connections in $T$ between $\mathcal{S}_i$ and $M$ evolve in a distinct subpopulation. At the same time new hidden nodes can be added to $M$; they evolve within their own subpopulations in the manner of standard ESP. In this way, the integrated evolutionary process simultaneously searches the space for how to reuse each potential source network and how to innovate with each new node. The GRUSM-ESP architecture (Figure~\ref{FigureGESP}) is composed of the following elements: (1) A pool of potential source networks. In the experiments in this paper, each target network reuses at most one source at a time; (2) \emph{Transfer genomes} encoding the weights of cross-network connections between source and target. Each potential source network in the pool has its own subpopulation for evolving transfer genomes between it and the target network. Each connection in $T$ is contained in some transfer genome. In our experiments, the transfer connections included are those such that the target's inputs are fully connected to the source's hidden layer, and the source's outputs are fully connected into the target's outputs; (3) A burst mechanism that determines when innovation is necessary based on a recent history of performance improvement. New hidden recruits (source networks when available, and single nodes otherwise) added during the burst phase evolve within their own subpopulations as in standard ESP.

All hidden and output neurons use a hyperbolic tangent activation function. Networks include a single hidden layer, and include recurrent self loops on hidden nodes; they are otherwise feedforward. The details of the genetic algorithm in our implementation used to evolve each subpopulation mirror those described by \citeauthor{GomezThesis} (\citeyear{GomezThesis}). This algorithm has been shown to work well within the ESP framework, though any suitable evolutionary algorithm could potentially be substituted in its place. (Preliminary experiments using this approach were discussed in \citeauthor{Braylan15b} (\citeyear{Braylan15b}).)

\section{Experiments}
\label{SectionExperiments}

GRUSM-ESP was evaluated in a stochastic version  of the Atari 2600 general video game-playing platform using the Arcade Learning Environment simulator (ALE; \citeauthor{Bellemare13} \citeyear{Bellemare13}). Atari 2600 is currently a very popular platform, because it challenges modern approaches, contains non-markovian games, and entertained a generation of human video game players, who would regularly reuse knowledge gained from previous games when playing new games. To make the simulator more closely resemble the human game-playing experience, the $\epsilon$-repeat action approach as suggested by \citeauthor{Hausknecht15} (\citeyear{Hausknecht15}) is used in this paper to make the environment stochastic; in this manner, like human players, the algorithm cannot as easily find loopholes in the deterministic nature of the simulator. The recommended $\epsilon=0.25$\footnote{\scriptsize{https://github.com/mgbellemare/Arcade-Learning-Environment/tree/dev}} is used. Note that the vast majority of previously published Atari 2600 results are in the deterministic setting; we are unaware of any existing scores that have been published in the $\epsilon$-repeat setting.

Agents were trained to play eight games: Pong, Breakout, Asterix, Bowling, Freeway, Boxing, Space Invaders, and Seaquest. Neuroevolution techniques are competitive in the Atari 2600 platform \cite{Hausknecht13}, and ESP in particular has yielded state-of-the-art performance for several games \cite{Braylan15a}. Three GRUSM-ESP conditions are evaluated: \texttt{\footnotesize{scratch}}, \texttt{\footnotesize{transfer}}, and \texttt{\footnotesize{random}}. In the \texttt{\footnotesize scratch} condition, networks are trained from scratch on a game using standard ESP (GRUSM-ESP with $S = \emptyset$). In the \texttt{\footnotesize transfer} condition, each scratch network is reused as a source network in training new GRUSM networks for different target games. In the \texttt{\footnotesize random} control condition, random networks are initialized and reused as source networks. Such networks contain on average the same number of parameters as fully-trained scratch networks.

Each run lasted 200 generations with 100 evaluations per generation. Since the environment is stochastic, each evaluation consists of five independent trials of individual $i$ playing game $g$, and the resulting score $s(i, g)$ is the average of the scores across these trials. The score of an evolutionary run at a given generation is the highest $s(i, g)$ achieved by an individual by that generation. A total of 333 runs were run split across all possible setups. Evolutionary parameters were selected based on their success with standard ESP.

To interface with ALE, the output layer of each network consists of a 3x3 substrate representing the nine directional movements of the Atari joystick in addition to a single node representing the Fire button. The input layer consisted of a series of object representations manually generated as previously described by \citeauthor{Hausknecht13} (\citeyear{Hausknecht13}). The location of each object on the screen was represented in an $8\times10$ input substrate corresponding to the object's class. The numbers of object classes varied between one and four. Although object representations were used in these experiments, pixel-level vision could also be learned from scratch below the neuroevolution process, e.g., via convolutional networks as was done by \citeauthor{Koutnik14} (\citeyear{Koutnik14}).\\


\noindent \textbf{Domain Characterization} Understanding \emph{when} transfer will be useful is important for any transfer learning approach. In many cases, attempting transfer can impede learning, leading to \emph{negative transfer}, when an approach is not able to successfully adapt knowledge from the source to the target domain. Negative transfer is a serious concern for many practitioners \cite{Taylor09,Pan10}. To help understand when GRUSM-ESP should be applied, it is useful to consider the diverse array of games within a unified descriptive framework. Biological neural reuse is generally thought to be most useful in transferring knowledge from simple behaviors to more complex, and the vast majority of previous computational approaches do exactly that. Thus, the characterization of games in this paper is grounded by a sense of relative complexity.

\begin{figure}[t]
\includegraphics[]{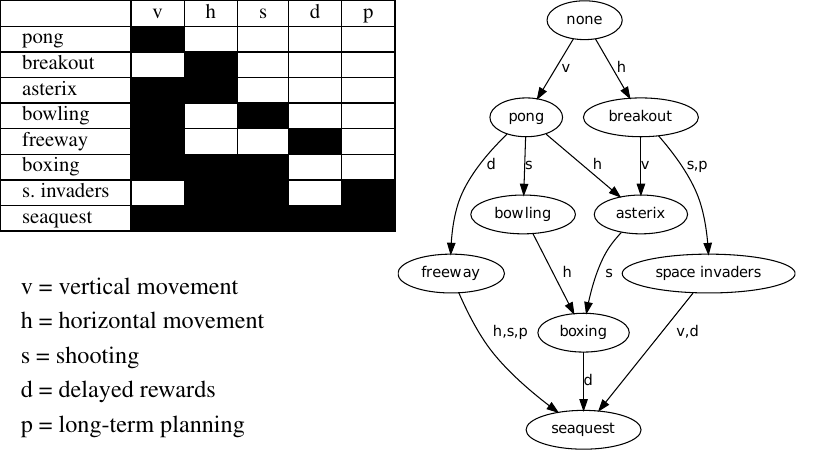}
\caption{\label{FigureFeatureLattice} (left) Feature representation for each game, and (right) games partially-ordered by feature inclusion: every path from \emph{none} to $g$ contains along its edges each complexity feature of $g$ exactly once, showing how games are related across the feature space. The existence of such a hierarchy motivates the use of atari for transfer.}
\end{figure}

Each game can be characterized by generic binary features that determine what successful game play requires: (1) horizontal movement (joystick left/right), (2) vertical movement (joystick up/down), (3) shooting (fire button); (4) delayed rewards; and (5) long-term planning. Intuitively, more complex games will include more of these features. A partial ordering of games by complexity defined by these features is shown in Figure~\ref{FigureFeatureLattice}. The assignment of features (1), (2) and (3) is completely defined based on game interface \cite{Bellemare13}. Freeway and Seaquest are said to have \emph{delayed rewards} because a high score can only be achieved by long sequences of rewardless behavior. Only Space Invaders and Seaquest were deemed to require long-term planning \cite{Mnih15}, since the long-range dynamics of these games penalize reflexive strategies, and as such, agents in these games can perform well with a low frequency decision-making \cite{Braylan15a}. In addition to being intuitive, these features are validated below based on how well they characterize games by complexity and how well they predict successful transfer.\\

\begin{figure}[t]
\centering{
\includegraphics[width=0.46\textwidth]{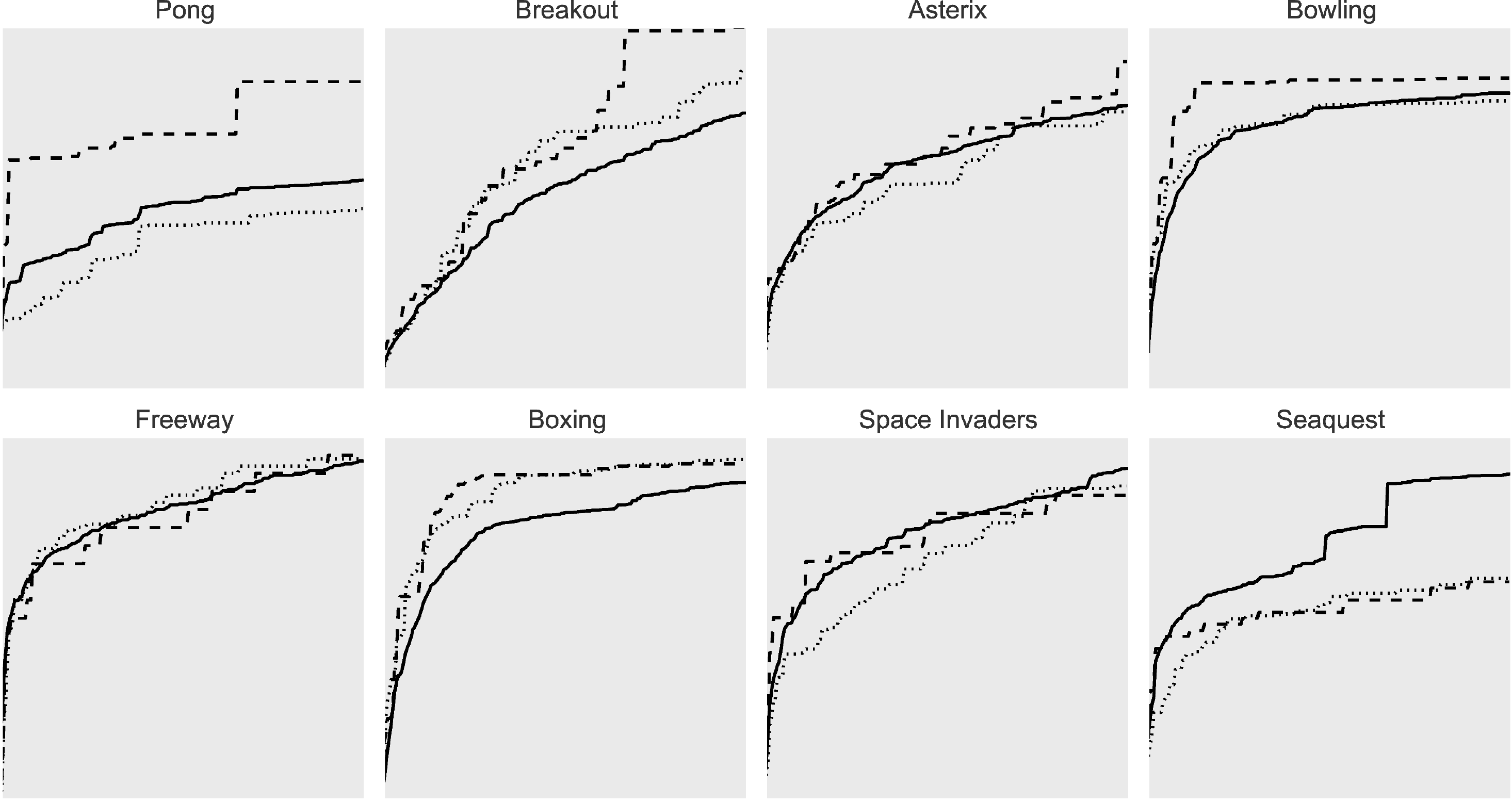}
}
\caption{\label{FigureLearningCurves} Raw mean score learning curves by generation for each target game aggregated over \texttt{\scriptsize transfer} (solid), \texttt{\scriptsize random} (dashed), and \texttt{\scriptsize scratch} (dotted) setups. The diversity of these learning curves shows the difficulty in comparing performance across games.}
\end{figure}

\noindent \textbf{Analysis Methods} There are many possible metrics for evaluating success of transfer, depending on what kind of transfer is desired or expected. Learning curves are irregular across different games, as illustrated in Figure~\ref{FigureLearningCurves}, which makes it difficult to choose a single metric that makes sense across all source-target pairs. Thus, the analysis is focused on a broad notion of \emph{transfer effectiveness} (TE), which aggregates metrics such as jumpstart and max overall score, with a weighted approximation of area under the curve \cite{Taylor09}. \emph{Success} of a setup is defined as the sum of the average score of that setup at a series of non-uniformly-spaced generations: $[1, 10, 50, 100, 200]$. This series favors early performance over later performance, as in general, in the long run, training from transfer and scratch should converge, as scratch eventually relearns everything that was effectively transferred. Then, the TE of a setup is its success minus the success of the control on the target game, the difference normalized by the size of the range of max scores achieved across all runs for that game, in order to draw comparison across games.

The first hypothesis is that \texttt{\footnotesize transfer} would outperform \texttt{\footnotesize scratch} in some setups, and that those setups could be predicted (i.e., they are not coincidental). However, any outperformance of \texttt{\footnotesize transfer} over \texttt{\footnotesize scratch} could be due to a larger number of network parameters. Therefore, as a second hypothesis, \texttt{\footnotesize random} setups were used as a control for the number of parameters, to test how \texttt{\footnotesize transfer} could predictably outperform \texttt{\footnotesize random}. We postulated and tested several useful indicators for predicting the outperformance of transfer, i.e., TE: (1) \emph{feature similarity}: count of features that are 1 for both source and target); (2) \emph{source feature complexity}: feature count of source game; (3) \emph{target feature complexity}: feature count of target game; (4) \emph{source training complexity}: source game average time to threshold; (5) \emph{target training complexity}: target game average time to threshold, where the threshold for each game is the minimum max score achieved across all \texttt{\footnotesize{scratch}} runs for that game, and time to threshold is the average number of generations to reach this threshold.

To predict TE, a linear regression model was trained in a leave-one-out cross-validation analysis. For each possible source-target pair $(s, t)$, the model was trained on all pairs $(s', t' \neq t)$ with TE as the dependent variable and the five indicators as the independent variables. Subsequently, the trained model was used to predict the TE of $(s, t)$. Correlation between the actual and predicted TE across all test pairs was used to gauge the predictability of TE. This experiment was conducted identically for both \texttt{\footnotesize transfer} versus \texttt{\footnotesize scratch} and \texttt{\footnotesize transfer} versus \texttt{\footnotesize random} conditions.

\subsubsection{Results}

\begin{figure}[t]
\centering{
\includegraphics[width=0.23\textwidth]{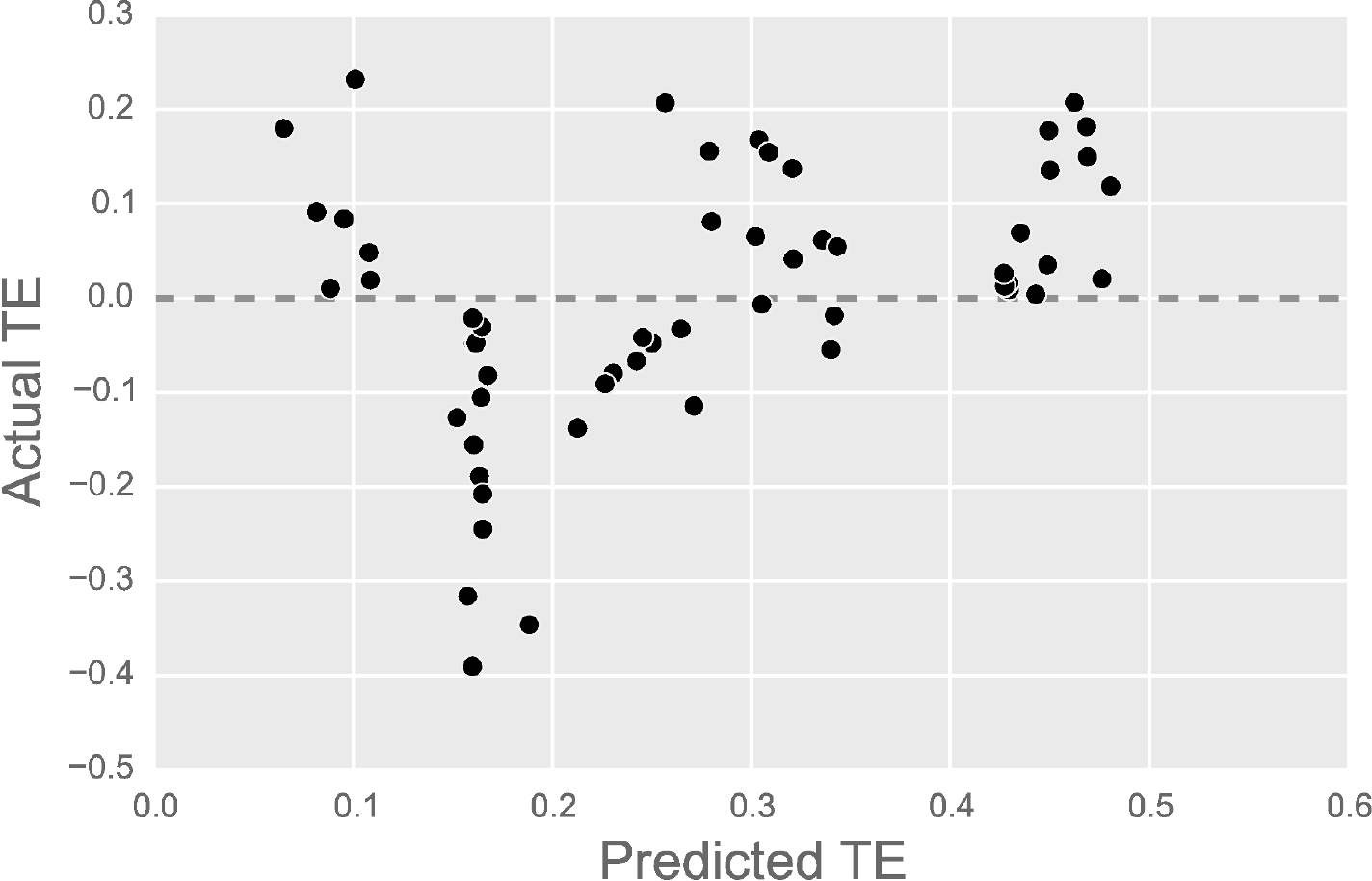}
\includegraphics[width=0.23\textwidth]{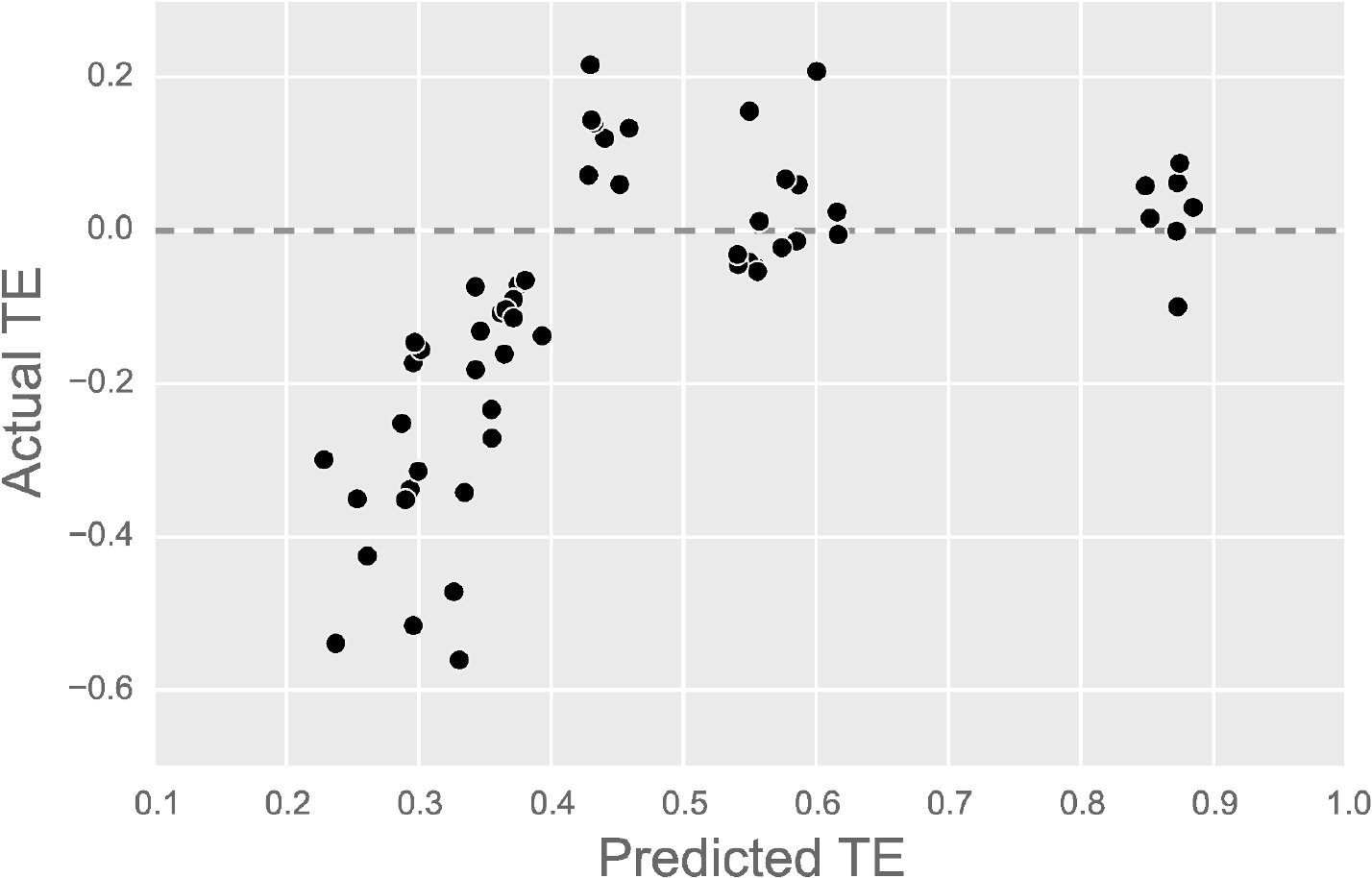}
}
\caption{\label{FigureScatterplots} Predicted vs. actual transfer effectiveness with respect to \texttt{\scriptsize scratch} (left) and \texttt{\scriptsize random} (right). Both predictors have a significant correlation between predicated and actual transfer effectiveness.}
\end{figure}

For both hypotheses, the indicator-based model proved to be a statistically significant predictor of transfer effectiveness in the test data: correlation $R = 0.40$ and p-value $< 0.0025$ for \texttt{\footnotesize transfer} versus \texttt{\footnotesize scratch}; correlation $R = 0.53$ and p-value $< 10^{-7}$ for \texttt{\footnotesize transfer} versus \texttt{\footnotesize random} (Figure~\ref{FigureScatterplots}). The strongest indicators for \texttt{\footnotesize transfer} versus \texttt{\footnotesize scratch} were target feature complexity and target training complexity, and for \texttt{\footnotesize transfer} versus \texttt{\footnotesize random} the strongest indicator was target feature complexity.

\begin{figure}[t]
\centering{
\includegraphics[width=0.45\textwidth]{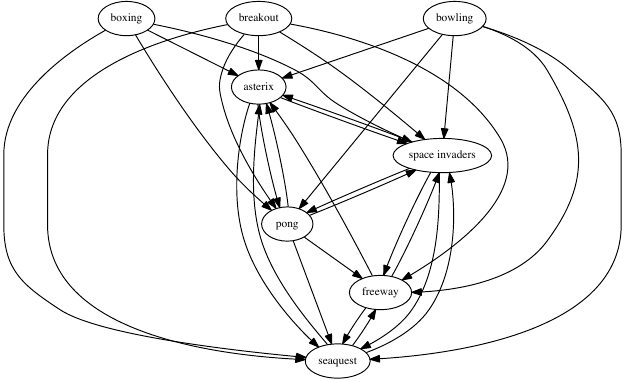}
\includegraphics[width=0.43\textwidth]{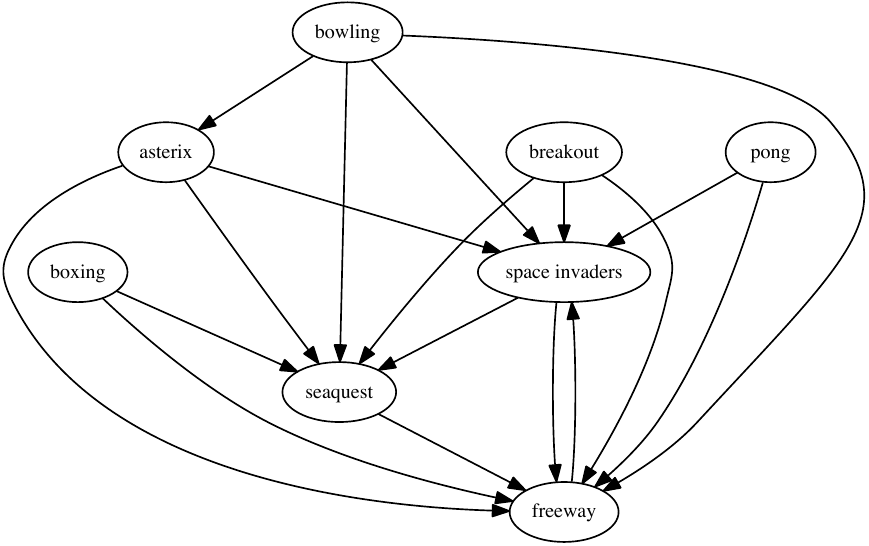}
}
\caption{\label{FigureTransferNetworks} Transferability graphs over all pairs of tasks with respect to \texttt{\scriptsize scratch} (top) and \texttt{\scriptsize random} (bottom) illustrating the target-centric clustering of successful source-target pairs. Each graph includes a directed edge from $g_1$ to $g_2$ $\iff$ the TE (see Analysis) for $g_2$ reusing $g_1$ is positive.}
\end{figure}

The fact that more complex games are more successful targets should not be surprising. As noted before, in most transfer learning scenarios, only simple-to-complex transfer is considered. The ability to predict when GRUSM-ESP will work is an important tool when applying this method to larger problems, and it is encouraging that the predictive indicator coincides with the `common sense' expectations of transfer effectiveness in the current experiments. TE for all source-target pairs is visualized in Figure~\ref{FigureTransferNetworks}. Also, although it is difficult to compare to the deterministic Atari 2600 domain, Table~\ref{TableComparison} provides a comparison of GRUSM-ESP to recent results in that domain for context \cite{Mnih15}.

\begin{table}
\scriptsize
\centering{
\begin{tabular}{|l|r|r|r|r|r|}
\hline
\rule{0pt}{3ex}game & \texttt{\scriptsize scratch} & \texttt{\scriptsize random} & \texttt{\scriptsize transfer} & human & DQN \\ \hline
\rule{0pt}{2ex}pong & 0.0 & 21.0 & $10.0_{ast}$ & 9.3 & 18.9 \\ \hline
\rule{0pt}{2ex}breakout & 31.0 & 35.0 & $30.3_{box}$ & 31.8 & 401.2 \\ \hline
\rule{0pt}{2ex}asterix & 2800 & 3216.7 & $3355_{bow}$ & 8503 & 6012 \\ \hline
\rule{0pt}{2ex}bowling & 249.3 & 265.0 & $265.0_{fr}$ & 154.8 & 42.4 \\ \hline
\rule{0pt}{2ex}freeway & 31.4 & 31.5 & $32.2_{brk}$ & 29.6 & 30.3 \\ \hline
\rule{0pt}{2ex}boxing & 93.9 & 92.7 & $95.0_{sea}$ & 4.3 & 71.8 \\ \hline
\rule{0pt}{2ex}s. invad. & 1438.0 & 1407.5 & $1655.0_{po}$ & 1652 & 1976 \\ \hline
\rule{0pt}{2ex}seaquest & 466.0 & 460.0 & $975.0_{sp}$ & 20182 & 5286 \\ \hline
\end{tabular}}
\caption{\label{TableComparison} For each game, average scores for \texttt{\scriptsize scratch}, \texttt{\scriptsize random}, and \texttt{\scriptsize transfer} from best source (subscripted). Interestingly, the best source for each target is unique. We also show human and DQN scores \cite{Mnih15}. Note: DQN uses deterministic ALE, so the most apt external comparison here may be to humans, who cannot deterministically optimize trajectories at the frame level.}
\end{table}

\section{Discussion and Future Work}
\label{SectionDiscussion}

The results show that GRUSM-ESP (an evolutionary algorithm for general transfer of neural network structure) can improve learning in Atari game playing by reusing previously developed knowledge. They also make it possible to characterize the conditions under which transfer may be useful. More specifically, the improvement in learning performance in the target domain depends heavily on the complexity of the target domain. The effectiveness of transfer in complex games aligns with the common-sense notion of hierarchical knowledge representation, as argued previously in transfer learning \cite{Konidaris12} as well as in biology \cite{Anderson10,Milo02}. It will be interesting to investigate whether the same principles extend to other general video game playing platforms, such as VGDL \cite{Perez15,Schaul13}. Such work should help understand how subsymbolic knowledge can be recycled indefinitely across diverse domains. 

Transfer is likely inefficient in simpler games due to the effort involved in finding the necessary connections for reusing knowledge from a given source network effectively, in which case it is more efficient to relearn from scratch. For particular low-complexity games, it can also be seen that \texttt{\footnotesize random} consistently outperforms both \texttt{\footnotesize scratch} and \texttt{\footnotesize transfer} (e.g., pong). The initial flexibility of untrained parameters in the \texttt{\footnotesize random} condition may explain this result. Unfreezing reused networks, and allowing them to change with a low learning rate may help close this gap.

Some \texttt{\footnotesize transfer} pairs do not consistently outperform training from \texttt{\footnotesize scratch} or \texttt{\footnotesize random}, indicating negative transfer. This highlights the importance of source and target selection in transfer learning. These results have taken a step towards answering the target-selection problem: What kinds of games make good targets for transfer? More data across many more games is required to answer the source-selection problem: For a given game, what sources should be used? A next step will involve pooling multiple candidate sources and testing GRUSM-ESP's ability to exploit the most useful structure available.

Despite negative transfer in some of the setups, the technique of training a classifier to predict transfer success is shown to be a useful approach for helping decide when to transfer: given some space of complex disparate domains, try transfer with a subset of source-target pairs, and use the results to build a classifier to inform when to attempt transfer in the future. In this paper, domain-characterization features were provided, but  domain-agnostic features could be learned from analysis of the networks and/or learning process; this is an interesting avenue for future work.

Another area of future work involves increasing the flexibility in the combined architecture by (1) relaxing the requirement for all transfer connections to be input-to-hidden and output-to-output, (2) allowing deeper architectures for the source and target networks, and (3) including multiple source networks with adaptive connectivity to each. These extensions will promote reuse of subnetworks of varying depth, along with flexible positioning and combination of modules. However, for GRUSM-ESP, as networks become large and plentiful, maintaining full connectivity between layers will become intractable, and it will be necessary to enforcing sparsity. GRUSM-ESP can also be extended to include LSTM units, e.g., as by \citeauthor{Schmidhuber07} (\citeyear{Schmidhuber07}), when deep memory is a primary concern.

\section{Conclusion}
\label{SectionConclusion}
This paper introduced an approach for general transfer learning using neural networks. The approach minimizes a priori assumptions of task relatedness and enables a flexible approach to adaptive learning across many domains. In a stochastic version of the Atari 2600 general video game-playing platform, a specific implementation developed in this paper as GRUSM-ESP can boost learning by reusing neural structure across disparate domains. The success of transfer is shown to correlate with intuitive notions of domain complexity. These results indicate the potential for general neural reuse to predictably assist agents in increasingly complex environments.\\

{\small \noindent \textbf{Acknowledgments} We would like to thank Ruohan Zhang for useful feedback. This research was supported in part by NSF grant DBI-0939454, NIH grant R01-GM105042, and an NPSC fellowship sponsored by NSA.}

\bibliographystyle{aaai}
{\small \bibliography{grusm}}

\end{document}